%% file: main.tex
\documentclass[letterpaper, 10pt, conference]{ieeeconf}
\IEEEoverridecommandlockouts
\input{preamble}

\title{\fontsize{17pt}{24pt}\selectfont \bf Sampling-based Kinodynamic Planning with World Models}

\newif\ifanonymous
\anonymousfalse

\ifanonymous
  \author{Anonymous Author(s)}
\else
    \author{
    Yitian Gao, Hanlan Yang, and Zachary Kingston%
    \thanks{YG, HY, and ZK are with the Department of Computer Science, Purdue University, {\tt \{gao634, yang3034, zkingston\}@purdue.edu}.
    }}
\fi

\begin{document}
\maketitle
\thispagestyle{empty}
\pagestyle{empty}

\setlength{\abovedisplayskip}{3pt} %
\setlength{\belowdisplayskip}{3pt} %
\setlength{\abovedisplayshortskip}{3pt} %
\setlength{\belowdisplayshortskip}{3pt} %
\setlength{\floatsep}{0pt}%
\setlength{\textfloatsep}{0pt}%
\setlength{\intextsep}{0pt} %
\setlength{\belowcaptionskip}{0.7em} %
\setlength{\dbltextfloatsep}{0pt} %
\setlength{\dblfloatsep}{0pt} %

\input{sections/00-abstract}
\input{sections/01-intro}

\input{sections/02-related}

\input{sections/03-prelim}

\input{sections/04-method}

\input{sections/05-experiment}

\input{sections/07-conclusion}

\input{sections/10-ack}

\printbibliography{}

\end{document}

%% file: preamble.tex
\usepackage[utf8]{inputenc}

\usepackage{comment}
\usepackage{xspace}
\usepackage{graphicx}
\usepackage{overpic}
\usepackage{svg}
\usepackage{xcolor}
\usepackage{bm}
\usepackage[hang,flushmargin]{footmisc}
\usepackage{adjustbox}

\usepackage{tabularx}
\usepackage{multirow}
\usepackage{booktabs}
\usepackage{colortbl}
\usepackage{float}
\usepackage{placeins}
\usepackage{xspace}

\usepackage{array}
\newcolumntype{P}[1]{>{\centering\arraybackslash}p{#1}}  %

\usepackage{etoolbox}
\makeatletter
\patchcmd{\@makecaption}
{\scshape}
{}
{}
{}
\newcommand{\onetagright}{\tagsleft@false}
\makeatother

\usepackage{amsmath}
\usepackage{amssymb}
\usepackage{amsthm}
\usepackage{siunitx}

\DeclareMathOperator*{\argmin}{arg\,min}

\newtheoremstyle{main}
{1em}                                                %
{1em}                                                %
{\itshape}                                        %
{0pt}                                                %
{\scshape}                                           %
{\\*}                                                %
{2pt}                                                %
{\thmname{#1}\thmnumber{ #2}: \thmnote{\itshape #3}} %

\usepackage{environ}

\usepackage[linesnumbered,ruled,noend]{algorithm2e}
\newcommand{\removelatexerror}{\let\@latex@error\@gobble}

\usepackage[
	activate   = {true},
	protrusion = false,
	expansion  = true,
	kerning    = true,
	spacing    = true,
	tracking   = false,
	auto       = true,
	selected   = true,
	factor     = 1000,
	stretch    = 10,
	shrink     = 10,
]{microtype}

\usepackage{csquotes}
\usepackage[
	maxbibnames=99,
	maxcitenames=2,
	natbib=true,
	bibstyle=ieee,
    citestyle=numeric-comp,
	backend=biber,
	sorting=none,
	giveninits=true,
	url=false,
	doi=false,
	eprint=false,
	isbn=false,
]{biblatex}

\definecolor{purduegold}{HTML}{C28E0E} %
\definecolor{figstart}{RGB}{42,150,215}
\definecolor{figgoal}{RGB}{34,160,34}
\definecolor{figreach}{RGB}{255,120,40}
\definecolor{figpath}{RGB}{250,60,60}

\makeatletter
\let\NAT@parse\undefined
\makeatother
\usepackage[pdfa,colorlinks,bookmarksopen,bookmarksnumbered,allcolors=black]{hyperref}
\usepackage[english]{babel}

\makeatletter
\newcommand\footnoteref[1]{\protected@xdef\@thefnmark{\ref{#1}}\@footnotemark}
\makeatother

\usepackage[nameinlink,capitalise]{cleveref}
\crefname{line}{line}{lines}
\crefname{figure}{Fig.}{Figs.}
\Crefname{figure}{Fig.}{Figs.}
\crefname{equation}{Eq.}{Eqs.}
\Crefname{equation}{Eq.}{Eqs.}
\crefname{section}{Sec.}{Secs.}
\Crefname{section}{Sec.}{Secs.}
\crefname{definition}{Def.}{Defs.}
\Crefname{definition}{Def.}{Defs.}
\crefname{algorithm}{Alg.}{Algs.}
\Crefname{algorithm}{Alg.}{Algs.}
\crefname{assumption}{Asm.}{Asms.}
\Crefname{assumption}{Asm.}{Asms.}
\crefname{subassumption}{Asm.}{Asms.}
\Crefname{subassumption}{Asm.}{Asms.}
\Crefname{problem}{Problem}{Problems}
\crefname{problem}{Problem}{Problems}

\definecolor{bestcell}{HTML}{DCE8F5}   %
\definecolor{secondcell}{HTML}{FDEAD6}  %
\providecommand{\rkfirst}[1]{\colorbox{bestcell}{\textbf{{#1}}}}
\providecommand{\rksecond}[1]{\colorbox{secondcell}{\underline{#1}}}

\usepackage{pbalance}
\usepackage{flushend}

\let\labelindent\relax
\usepackage[inline]{enumitem}
\usepackage{mathtools}
\usepackage[normalem]{ulem}

\graphicspath{ {./images/} }

\newcommand{\methodname}{RRT\xspace}           %
\newcommand{\cem}{CEM\xspace}
\newcommand{\mppi}{MPPI\xspace}

\usepackage{subcaption}
\usepackage{threeparttable}

\usepackage{newtxtext,newtxmath}

%% file: sections/00-abstract.tex
\begin{abstract}

Visual world models predict the consequences of actions from images, but planning with them is typically done by optimizers such as CEM and MPPI.
However, these optimizers stall on long-horizon tasks where progress requires motion the objective does not reward, such as moving around a block before pushing it.
This paper revisits kinodynamic sampling-based motion planning for world models, using the learned model as the propagator of a tree search in latent space.
Our key insight is that the sampling strategies devised for high-dimensional search in configuration space apply directly to the latent space of a world model, where they rapidly expand through intermediate states that myopic optimizers never sample.
We present two exploration strategies for the high-dimensional latent space, nearest-neighbor-guided RRT and projection-based KPIECE, with GPU-native batched expansion for efficient device utilization.
We evaluate with an off-the-shelf DINO-WM on four simulated domains and a real-world block-pushing task.
Our approaches match or exceed the success rate of CEM and MPPI at up to $13.8\times$ lower planning time in simulation, and improve the success rate by $20\%$ on the real robot.

\end{abstract}

%% file: sections/01-intro.tex
\section{Introduction}
\label{sec:intro}

Visual world models learn to predict the consequences of actions from images~\cite{worldmodels}.
Given a history of observations and a candidate action, the model predicts the next observation or latent state, and this prediction can stand in for a simulator when no analytical model of the task is available~\cite{planet, dreamerv3, dinowm, vjepa2, cosmos}.
For example, a robot pushing a block across a table has no exact closed-form model of the contact, but a world model trained on demonstrations can predict where the block will end up (\cref{fig:live-frames}b).
The remaining question is how to search through the model's predictions to find actions that reach a goal.

Prior works~\cite{pets, planet, tdmpc2, dinowm, vjepa2} typically use closed-loop planning, i.e., they repeatedly generate plans with the Cross-Entropy Method (CEM)~\cite{cem} or Model-Predictive Path Integral control (MPPI)~\cite{mppi}, then execute action prefixes with receding-horizon Model Predictive Control (MPC)~\cite{mpc}. However, gradient-free methods like CEM and MPPI require repeated rollouts for decision-making. Combined with costly inference on large-scale world models, this requires care to balance planning quality and computational cost. Additionally, CEM and MPPI perturb around a mean, leading them to focus exploration and become trapped in local minima. Variants reshape the sampling distribution~\cite{icem, biasedmppi} but keep the same search structure. With exact dynamics, tree search under a receding horizon has been shown to outperform CEM~\cite{mpt}; others have proposed to use search methods such as Monte Carlo tree search~\cite{worldplanner}, but rely on an action prior. These gaps raise the question, are optimization-based approaches the right tools for exploring and planning in the latent space of world models?

On the other hand, classical search-based and sampling-based motion planning methods (SBMPs) excel in long-horizon planning for robotics~\cite{lavalle2006planning}, as they maintain a tree or graph that guides exploration. However, they have not been widely adopted as planning methods with world models due to the curse of dimensionality. A large body of literature tackles this problem in robotics with more efficient sampling and exploration strategies designed for high-dimensional spaces~\cite{syclop, kpiece, bitstar}; others use parallel hardware to speed up the planning process~\cite{cforest, aepase, pachs, prrtc}. These advances motivate bridging the aforementioned research gaps in world model planning with classical planning methods.

Thus, we revisit kinodynamic SBMP for planning with world models.
Our key insight is that the sampling strategies devised for high-dimensional search in robot configuration space apply directly to the latent space of a world model.
Nearest-neighbor guidance (i.e., RRT~\cite{rrt}) and projection-based coverage estimation (i.e., KPIECE~\cite{kpiece}) need only a compact descriptor of each state, while the world model propagates the full latent along each branch.
Batching propagation across candidate expansions amortizes the predictor's cost, allowing our planner to match or exceed the success rate of CEM and MPPI at a fraction of their planning time, with a frozen, off-the-shelf world model. In this work, we present:

\begin{enumerate}[leftmargin=*, itemsep=2pt, topsep=2pt]
  \item a kinodynamic tree planner that uses a visual world model as its propagator, with batched expansion;
  \item two exploration strategies for the world model's latent space, nearest-neighbor-guided RRT and projection-based coverage guidance with KPIECE~\cite{kpiece};
  \item and an evaluation against CEM, MPPI, and MCTS~\cite{worldplanner} with DINO-WM on four simulated domains and a real-world block-pushing task, with ablations of exploration, refinement, and numerical precision.
\end{enumerate}

Across the simulated domains, our planners match or exceed the success rate of CEM and MPPI with up to \(13.8\times\) lower median planning time and an order of magnitude fewer world-model evaluations.
On the real robot, our planner solves \(29/30\) problems where CEM and MPPI solve \(23/30\), and the gap is widest on tasks that require a detour before pushing, \(9/10\) versus \(5/10\).

%% file: sections/02-related.tex
\section{Related Work}
\label{sec:lit}

Early work on visual prediction, such as Visual Foresight~\cite{Dvisforesight, visforesight}, combined action-conditioned video prediction with model-predictive control (MPC) and demonstrated that a data-driven visual predictive model can guide control toward goals in manipulation tasks. Other recent work advances visual world models through large-scale pretraining of visual encoders or predictive models. iVideoGPT~\cite{ivideogpt} pretrains an autoregressive model on human and robot manipulation trajectories and adapts it to action-conditioned prediction, visual planning, and model-based reinforcement learning. DINO-WM~\cite{dinowm} learns action-conditioned dynamics over frozen DINOv2~\cite{dinov2} patch features, while V-JEPA2~\cite{vjepa2} builds on pretrained video representations and demonstrates image-goal planning on physical robots. However, due to the fidelity gap, these world models typically use CEM to optimize action sequences within a receding-horizon MPC framework. WorldPlanner and STORM~\cite{worldplanner, storm} use Monte Carlo Tree Search (MCTS) to guide planning.

Learned components have been integrated into SBMP at every stage, including samplers, steering, distance metrics, collision checking, and action proposals inside a kinodynamic tree~\cite{mlsbmpsurvey, ditree}. In each case the learned component operates in a classical state space and the propagator remains a simulator or analytic model. One line of work instead plans in a learned latent. L2RRT~\cite{l2rrt} learns a compact state representation and a collision classifier, then constructs a search tree through sampled controls and learned forward propagation through latent dynamics. ELAST~\cite{ELAST} extends this direction through exploration in learned visual representations, incorporating model uncertainty and transition density. VELAP~\cite{velap} adds RRT-style expansion and prunes transitions by Q-ensemble disagreement. These methods train their latent on task data and construct compact representations for planning, yet sampling-based kinodynamic planning with the larger patch-feature latent representations used in contemporary visual world models remains underexplored.

Our approach draws inspiration from L2RRT and KPIECE~\cite{l2rrt, kpiece}. While sharing the fundamental principle of constructing reachable trajectories through learned forward propagation, we focus on integrating established SBMP methods with existing visual predictors whose rich representations, temporal context, and inference costs impose different computational demands. Through this integration, we examine the practical value of established kinodynamic planning techniques for contemporary visual world models on long-horizon robotic tasks including contact-mediated manipulation.

%% file: sections/03-prelim.tex
\section{Preliminaries}
\label{sec:prelim}

\subsection{Kinodynamic Motion Planning}

Consider a robot with state space \(\mathcal{X} \subseteq \mathbb{R}^n\) and control space
\(\mathcal{U} \subseteq \mathbb{R}^m\). The system dynamics are governed by
\begin{equation}
  \dot{x}(t) = f(x(t),u(t)),
  \label{eq:dynamics}
\end{equation}
where \(x(t) \in \mathcal{X}\) and \(u(t) \in \mathcal{U}\) are the state and
control at time \(t\). A trajectory \(\sigma:[0,T]\rightarrow\mathcal{X}\) is
feasible if there exists a control trajectory
\(u:[0,T]\rightarrow\mathcal{U}\) such that
\(\dot{\sigma}(t)=f(\sigma(t),u(t))\) for all \(t\in[0,T]\).

Given an initial state \(x_{init}\in\mathcal{X}\) and a goal region
\(\mathcal{X}_{goal}\subseteq\mathcal{X}\), a solution is a feasible
trajectory satisfying
\begin{equation}
  \sigma(0)=x_{init}, \qquad
  \sigma(T)\in\mathcal{X}_{goal}.
\end{equation}
We plan in discrete time with a fixed step $\Delta \mathbf{t}$: a control $u_t$ is applied over $[t\Delta \mathbf{t},(t+1)\Delta \mathbf{t})$ and $x_{t+1}$ is the state \cref{eq:dynamics} produces from $x_t$.
A solution is then a control sequence $u_{0:T-1}$.

\begin{figure*}[ht]
  \centering
  \includegraphics[width=\textwidth]{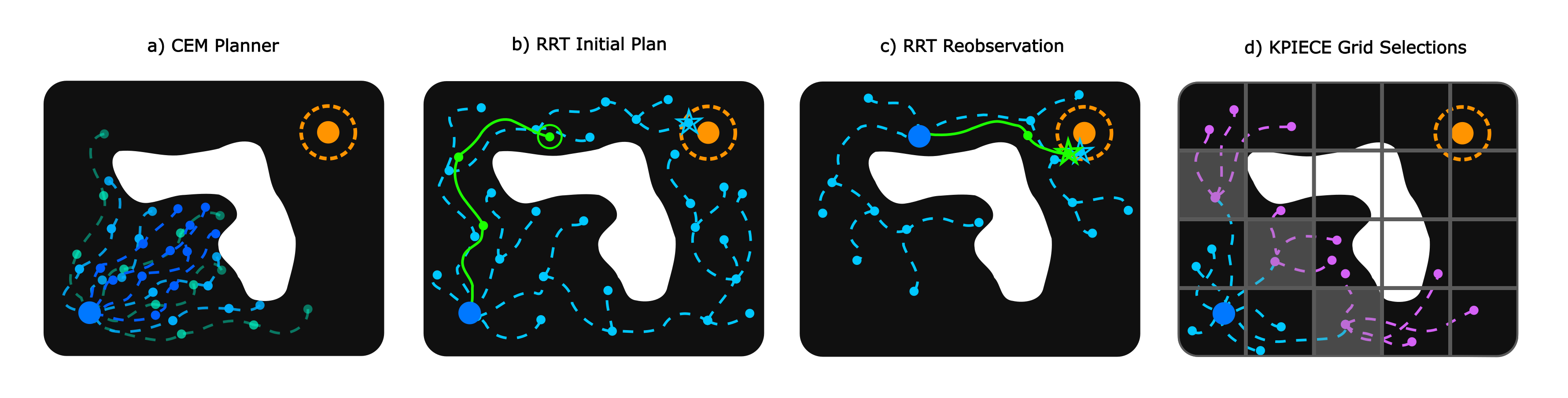}
  \caption{\textbf{Planning paradigms and receding-horizon execution in latent space.}
    (a) Shooting methods (CEM) iteratively collapse trajectories toward a mean, risking entrapment in local minima around non-convex obstacles.
    (b) Tree search (RRT) finds a valid path through the latent space, committing only its first \(n_{\mathrm{exec}}\) actions.
    (c) System execution introduces minor drift from the planned path, requiring re-grounding and replanning from the new observation.
  (d) KPIECE projects states onto a grid and prioritizes under-explored frontier cells (shaded) for subsequent expansions.}
  \label{fig:method-mpc}
\end{figure*}

\subsection{Learned Transition Function}
\label{sec:prelim:surrogate}

We consider settings where the true dynamics \(f\) and a simulator are
unavailable. Instead, a world model is learned from observation--action
sequences
\begin{equation}
  \mathcal{D}=\{(o_0,a_0,o_1,a_1,\ldots)\},
\end{equation}
where \(o_t\) denotes an observation of the system (which may include visual frames and proprioceptive state) and \(a_t\) the corresponding control.
The world model provides a learned transition function that predicts the next state representation \(z_{t+1}\) from a history of representations and actions,
\begin{equation}
  \hat{z}_{t+1} = g_\theta(z_{t-k+1:t}, a_{t-k+1:t}),
  \label{eq:predictor}
\end{equation}
where \(z_t \in \mathcal{Z}\) is the compact latent state representation. 

\subsection{Planning in Latent Space}
\label{sec:prelim:latent_planning}

In this work, planning in latent space is realized by using the learned transition function \(g_\theta\) as a forward propagator in place of true system dynamics or a physics engine.
A goal observation \(o_{goal}\) is represented by \(z_{goal}\), and the planner
considers the goal reached when
\begin{equation}
  d(z_t,z_{goal}) < \varepsilon.
\end{equation}
The classical formulation adds a validity constraint \(\sigma(t)\in\mathcal{X}_{free}\) for a collision-free set \(\mathcal{X}_{free}\subseteq\mathcal{X}\).
The learned model provides no such check, as obstacles are implicit in the transition model, which predicts no motion into them.
An alternative is to learn a separate classifier for where the model can be trusted~\cite{mitrano2021trust}; we instead rely on the transition model alone and correct its errors through replanning.

The latent goal criterion guides planning, but final performance is evaluated in the true environment under task-specific success metrics. Consequently, reaching the goal in the learned surrogate does not guarantee success on the ground-truth system.

%% file: sections/04-method.tex
\section{Methodology}
\label{sec:meth}

Our approach embeds a kinodynamic RRT directly within the latent space of an off-the-shelf world model.
The planner operates in a receding-horizon framework, repeatedly generating a plan from the latest observation and executing a short prefix.
We first detail the world-model representation and the receding-horizon loop, followed by our batched latent tree search and trajectory polishing.

\subsection{World-Model Planning}
\label{sec:meth:wm}

We use DINO-WM~\cite{dinowm} as our off-the-shelf world model.
The model is image-based, using a frozen DINOv2 encoder to map raw visual observations into latent features, which are combined with proprioceptive state into a unified latent state \(z_t \in \mathcal{Z}\).
The learned transition function \(g_\theta\) predicts the next state given action \(a_t\) and a context window of \(k=3\) latent frames (\(z_{t-2:t}\)).
Because \(g_\theta\) is trained on video demonstration data, each action step \(a_t\) represents a fixed temporal duration (\(\texttt{frameskip} = 5\)), discretizing the underlying continuous dynamics so transitions produce distinct latent changes.
For sampling-based planning, the choice of action distribution is critical.
We use the action prior provided by DINO-WM, which focuses the search on structured, temporally correlated actions likely to produce useful rollouts.

\subsection{Receding-Horizon Planning}
\label{sec:meth:mpc}

The planner operates within a receding-horizon framework, sketched in \cref{fig:method-mpc} and summarized in \cref{alg:mpc}.
At each iteration (Lines~\ref{alg:mpc:encode}--\ref{alg:mpc:plan}), the encoder constructs the current latent \(z_t\), and a planning method \(\mathcal{P}\) produces an action sequence \(a_{1:H}\) toward \(z_{goal}\) using \(g_\theta\) as its sole dynamic model.
While shooting-based methods like CEM gradually collapse trajectories toward a mean, they are vulnerable to local minima when progress requires first increasing the goal distance, as when the pusher must move around the block before pushing it (\cref{fig:method-mpc}a).
In contrast, tree-based planners explore around barriers to identify a feasible path to the goal (\cref{fig:method-mpc}b).
Only the first \(n_{\mathrm{exec}}\) world-model action steps of the resulting plan are executed on the real system before re-observing (Line~\ref{alg:mpc:exec}).
Because execution on the real system naturally deviates from the predicted trajectory (\cref{fig:method-mpc}c), each planning iteration begins fresh from the newly observed state.
We do not retain search trees between iterations or warm-start the trajectory optimizers, which prevents the accumulation of model error over long horizons.

\begin{algorithm}[t]
  \caption{Receding-horizon execution}
  \label{alg:mpc}
  \KwIn{History \(o_{t-k+1:t}\), goal \(o_{goal}\), encoder \(\phi\), propagator \(g_\theta\), planner \(\mathcal{P}\), horizon \(H\), prefix \(n_{\mathrm{exec}}\)}
  \KwOut{Actions executed on the real system}

  \While{task not complete}{
    \(z_{t-k+1:t} \leftarrow \phi(o_{t-k+1:t})\) \label{alg:mpc:encode}\;
    \(z_{goal} \leftarrow \phi(o_{goal})\)\;
    \(a_{1:H} \leftarrow \mathcal{P}\!\left(z_{t-k+1:t},\,z_{goal};\,g_\theta\right)\) \label{alg:mpc:plan}\;
    Execute \(a_{1:n_{\mathrm{exec}}}\) on the real system \label{alg:mpc:exec}\;
    \(o_{t-k+1:t} \leftarrow \textsc{Observe}()\)\;
  }
\end{algorithm}

\subsection{Tree Search in Latent Space}
\label{sec:meth:rrt}

Both tree planners share a single growth procedure, given in \cref{alg:tree}.
To conserve memory, we do not store the full latent history at every node; instead, the required context window is reconstructed from a node's ancestors immediately before propagation (Line~\ref{alg:tree:prop}).
Search typically terminates upon finding the first valid solution (Line~\ref{alg:tree:term}) or when the node budget is exhausted.
Node selection uses a goal bias.

\begin{algorithm}[t]
  \caption{Tree search in the latent surrogate}
  \label{alg:tree}
  \KwIn{Latents \(z_{init}, z_{goal} \in \mathcal{Z}\), propagator \(g_\theta\), prior \(\pi\), \textsc{Select}, bias \(\beta\), budget \(N\), tolerance \(\varepsilon\), polish \(m\)}
  \KwOut{Action sequence \(a_{1:H}\)}

  \(\mathcal{T} \leftarrow \{\textsc{Root}(z_{init})\}\)\;
  \While{\(|\mathcal{T}| < N\)}{
    \tcp{See \cref{alg:rrtsel} or \cref{alg:kpiece}}
    \(n \leftarrow \textsc{Select}(\mathcal{T},z_{goal},\beta)\) \label{alg:tree:select}\;
    \(a \sim \pi\)\;
    \(\hat{z} \leftarrow g_\theta(\textsc{History}(\mathcal{T},n),a)\) \label{alg:tree:prop}\;
    \(d \leftarrow d(\hat{z},z_{goal})\)\;
    \(\mathcal{T} \leftarrow \mathcal{T}\cup\{(\hat{z},n,a,d)\}\)\;
    \lIf{\(d<\varepsilon\)}{\textbf{break} \label{alg:tree:term}}
  }
  \(n^\star \leftarrow \argmin_{n\in\mathcal{T}} d(n)\)\;
  \(a_{1:H} \leftarrow \textsc{ExtractActions}(\mathcal{T},n^\star)\)\;
  \If{\(m>0\) \label{alg:tree:polish_start}}{
    \tcp{See \cref{sec:meth:heuristics}}
    \(a_{1:H} \leftarrow \textsc{PolishCEM}(a_{1:H},z_{init},z_{goal};g_\theta,m)\) \label{alg:tree:polish_end}\;
  }
  \Return \(a_{1:H}\)\;
\end{algorithm}

\begin{algorithm}[t]
  \caption{RRT selection rule (\textsc{Select} in \cref{alg:tree})}
  \label{alg:rrtsel}
  \KwIn{Tree \(\mathcal{T}\), goal latent \(z_{goal}\), goal bias \(\beta\)}
  \KwOut{Node to expand}

  \(u \sim \mathcal{U}[0,1]\)\;
  \lIf{\(u < \beta\)}{\(z_{rand} \leftarrow z_{goal}\)}
  \lElse{\(z_{rand} \sim \mathcal{Z}_{\mathrm{expl}}\)}
  \Return \(\argmin_{n\in\mathcal{T}} d(z_n,z_{rand})\)\;
\end{algorithm}
We batch node expansion and propagation, akin to PAKR~\cite{gao2026pakr}.
However, unlike purely analytic simulators, our maximum batch size is bottlenecked by the memory footprint of the world model's neural network propagator rather than the search algorithm itself.
Prior kinodynamic planners with expensive propagators have reduced cost by deferring propagation until an edge is needed~\cite{lazyboe}; we instead batch propagation so each call is amortized across candidates.

\subsection{Trajectory Polishing}
\label{sec:meth:heuristics}

A tree can only return an action sequence it explicitly sampled during expansion, meaning its commit is an \(\argmin\) over a discrete set.
We therefore apply a short CEM optimization to the extracted sequence before it is returned, shown in Lines~\ref{alg:tree:polish_start}--\ref{alg:tree:polish_end} of \cref{alg:tree}.
This polishing stage is initialized from the selected trajectory and optimizes it using the same world model and latent goal metric, recovering the continuous refinement the discrete search cannot express.

\subsection{Exploration Guidance in Latent Space}
\label{sec:meth:kpiece}
We provide two node-selection schemes to guide exploration in latent space. The first uses nearest-neighbor selection to induce Voronoi-biased exploration, as shown in \cref{alg:rrtsel}. A target state \(z_{\mathrm{rand}}\) is sampled, and the nearest tree node is selected for expansion. Under uniform target sampling, nodes with larger Voronoi regions are more likely to be selected, encouraging expansion toward sparsely explored regions. However, the effectiveness of this guidance depends on the distance function: in high-dimensional world-model representations, Euclidean distance may poorly reflect task-relevant differences between states.
We instead average the channel dimension across local spatial regions to produce a low-dimensional spatial feature map.
The nearest-neighbor distance $d(z_a, z_b)$ is then evaluated as the $L_2$ norm over these pooled spatial representations.

The second technique uses projection-guided exploration based on KPIECE~\cite{kpiece}, as shown in \cref{alg:kpiece}. It projects the world model's latent states into a low-dimensional exploration space, partitions this space into a grid, and prioritizes expansion from sparsely covered frontier cells (\cref{fig:method-mpc}d). This approach guides exploration through explicit coverage estimates in the low-dimensional space. We examine several projection methods, including random linear projections, principal component analysis (PCA), and spatial-coordinate projections such as \(XY\) position.

\begin{algorithm}[t]
  \caption{KPIECE selection rule (\textsc{Select} in \cref{alg:tree})}
  \label{alg:kpiece}
  \KwIn{Tree \(\mathcal{T}\), goal \(z_{goal}\), bias \(\beta\), projection \(\psi:\mathcal{Z}\rightarrow\mathbb{R}^{d_{\mathrm{proj}}}\), grid \(\mathcal{C}\)}
  \KwOut{Node to expand}

  \ForEach{node \(n\) added since the last call}{
    \(c \leftarrow \textsc{Cell}(\psi(z_n))\)\;
    \tcp{Update coverage of \(c\) and its neighbours}
    Insert \(n\) into \(c\) in \(\mathcal{C}\)\;
  }
  \(u \sim \mathcal{U}[0,1]\)\;
  \lIf{\(u < \beta\)}{\Return \(\argmin_{n\in\mathcal{T}} d(z_n,z_{goal})\)}

  \(c^\star \leftarrow \textsc{SelectCell}(\mathcal{C})\) \label{alg:kpiece:cell_sel}\;
  \Return \(\textsc{SelectState}(c^\star)\) \label{alg:kpiece:state_sel}\;
\end{algorithm}

The projection is used only for exploration and does not replace the latent state used by the world model.
Thus, KPIECE reduces the dimensionality of the exploration criterion while retaining the full representation for prediction and goal evaluation.

%% file: sections/05-experiment.tex
\begin{figure}[t]
  \centering
  \includegraphics[width=\columnwidth]{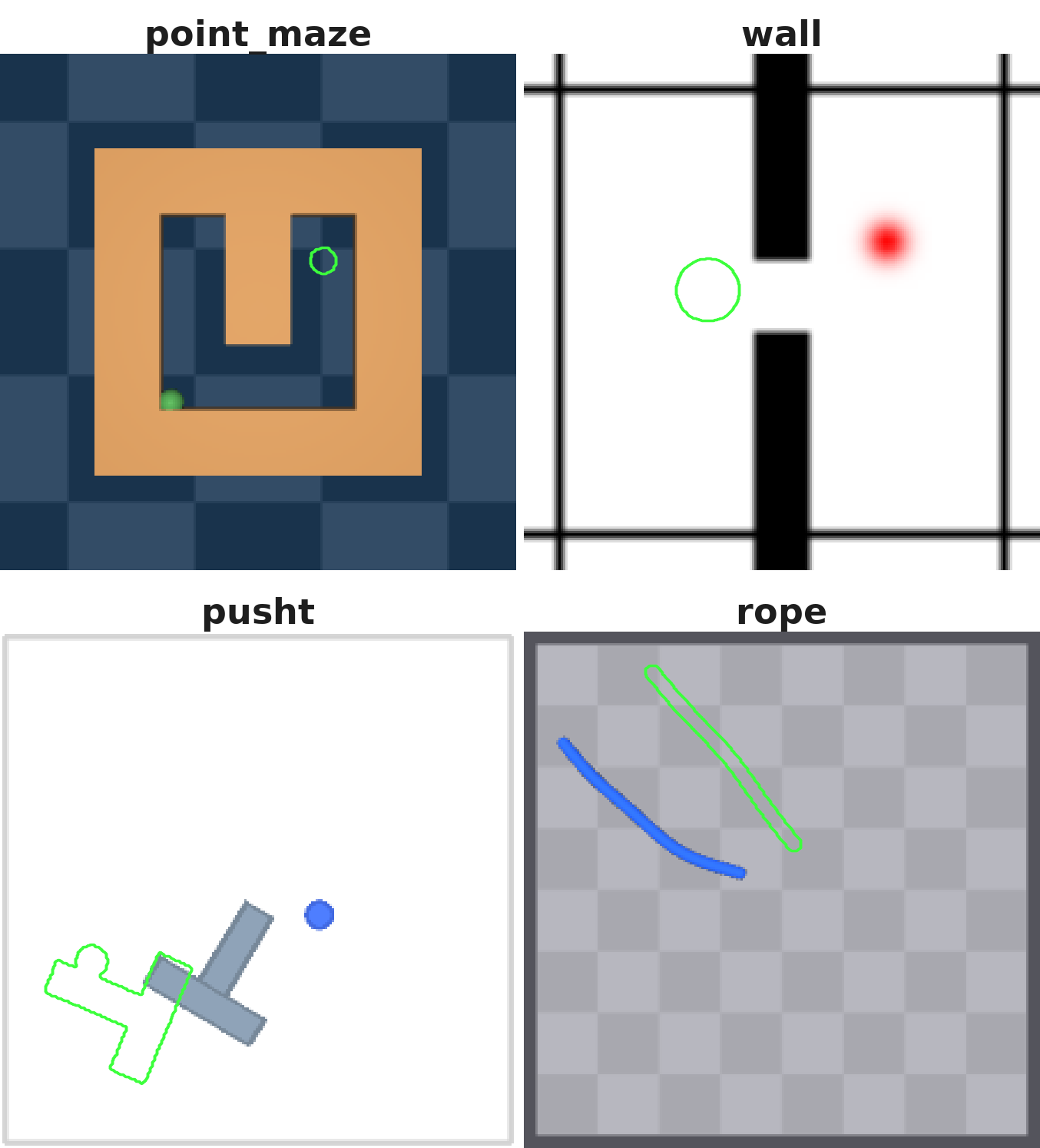}
  \caption{\textbf{The four simulation environments}. The start image of a representative problem from each environment is overlaid with a green goal position.}
  \label{fig:sim-envs}
\end{figure}
\begin{figure*}[ht]
  \centering
  \includegraphics[width=\textwidth]{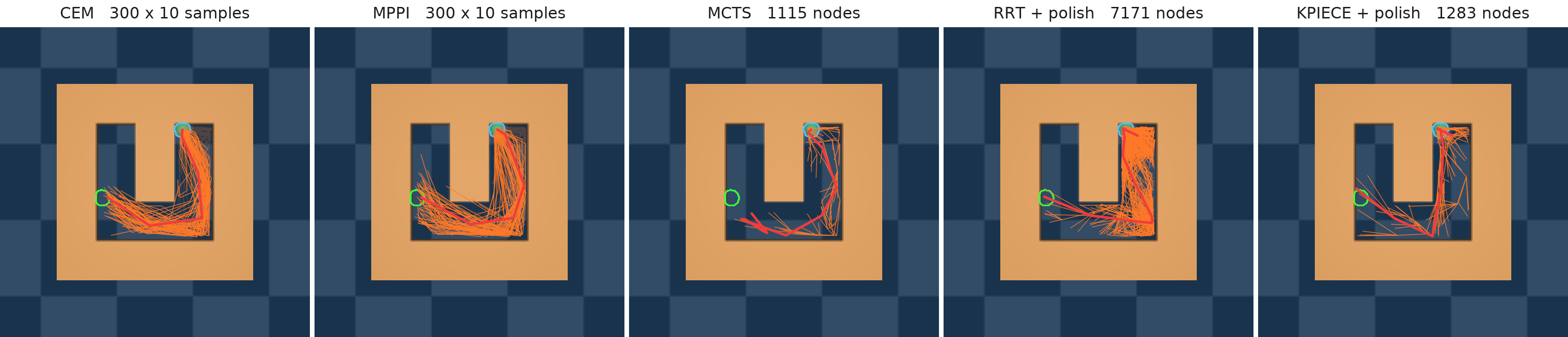}
  \caption{\textbf{Qualitative comparison of search exploration on \texttt{point\_maze}.} The initial plan of the same problem (start in blue, goal outlined in green). The depicted trees are thinned out to be $\sim$5x sparser. MCTS, RRT, and KPIECE reach the goal in the first plan, while CEM and MPPI execute their full budget.}
  \label{fig:sim-trees}
\end{figure*}

\section{Experiments}
\label{sec:exp}

All experiments are conducted on a virtual machine equipped with an NVIDIA RTX 4090 GPU with 24 GB of VRAM, 64 GB of system memory, and 12 CPU cores allocated from a 24-core AMD Ryzen Threadripper PRO 5965WX processor operating at 3.8 GHz.
The underlying world model uses the frozen DINOv2 encoder described in \cref{sec:meth:wm}, which remains untouched across all tests without planner access to ground-truth simulation dynamics during search.
Following the evaluation protocol established by DINO-WM~\cite{dinowm}, problem instances are defined by sampling fixed-horizon demonstration clips (\(H \in \{5, 10\}\) steps) and setting their sequence endpoints as start and goal configurations.
To maintain direct alignment with this standard benchmark, we evaluate across a screened random subset of these official task sets, filtering out trivial near-goal queries and demonstrator-failed trajectories.

\begin{table*}[t]
  \centering
  \caption{\textbf{Main comparison.} Evaluated over \(N{=}50\) problems per domain across identical world-model checkpoints and MPC settings. \emph{Time} shows median wall-clock time with mean in parentheses. \emph{Cost} indicates median solution cost. \emph{Nodes} denotes median world-model single-step evaluations for a whole problem, summed over its replans (inserted tree nodes for tree methods; \(\text{samples} \times \text{horizon} \times \text{optimization steps taken}\) for \cem{} and \mppi{}); it is therefore a multiple of the per-replan budgets in \cref{tab:budget} whenever the median solved problem needs more than one replan. \emph{Time}, \emph{Cost} and \emph{Nodes} are all computed exclusively over solved trials. Every arm evaluates the world model in \texttt{fp16} (\cref{tab:precision}), so the wall-clock column compares planners rather than numerics.}
  \label{tab:main}
  \small
  \setlength{\tabcolsep}{5pt}
  \begin{tabular}{@{}l cccc cccc@{}}
    \toprule
    & \multicolumn{4}{c}{\textbf{point\_maze} (\(H{=}10\))}
    & \multicolumn{4}{c}{\textbf{wall} (\(H{=}5\))} \\
    \cmidrule(lr){2-5} \cmidrule(lr){6-9}
    \textbf{Planner} & Succ.\(~\uparrow\) & Time\,s\(~\downarrow\) & Cost\(~\downarrow\) & Nodes\(~\downarrow\)
    & Succ.\(~\uparrow\) & Time\,s\(~\downarrow\) & Cost\(~\downarrow\) & Nodes\(~\downarrow\) \\
    \midrule
    \rowcolor{gray!10}
    \methodname (ours) & \rksecond{49/50} & \rksecond{0.61 (1.55)} & 5.0 & 2\,307 & 39/50 & \rksecond{0.26 (0.77)} & 3.0 & 770 \\
    \rowcolor{gray!10}
    KPIECE (ours) & 48/50 & \rkfirst{0.53 (1.24)} & 5.0 & 1\,412 & 42/50 & \rkfirst{0.20 (0.41)} & 3.0 & 769 \\
    \rowcolor{gray!10}
    \methodname + polish (ours) & \rkfirst{50/50} & 2.88 (4.41) & 5.5 & 2\,819 & 43/50 & 0.56 (0.80) & 3.0 & 769 \\
    \rowcolor{gray!10}
    KPIECE + polish (ours) & \rksecond{49/50} & 2.36 (3.84) & 4.0 & 1\,283 & \rkfirst{45/50} & 0.42 (1.14) & 2.0 & 513 \\
    \midrule
    MCTS~\cite{worldplanner} & 48/50 & 9.95 (10.72) & 8.0 & 1\,276 & 43/50 & 1.18 (1.02) & 5.0 & 384 \\
    \cem~\cite{dinowm} & \rkfirst{50/50} & 16.07 (20.20) & 5.0 & 45\,000 & \rksecond{44/50} & 1.61 (1.68) & 5.0 & 15\,000 \\
    \mppi & \rksecond{49/50} & 4.62 (6.56) & 5.0 & 15\,000 & 36/50 & 5.07 (5.25) & 5.0 & 45\,000 \\

    \midrule[\heavyrulewidth]
    & \multicolumn{4}{c}{\textbf{pusht} (\(H{=}5\))}
    & \multicolumn{4}{c}{\textbf{rope} (\(H{=}5\))} \\
    \cmidrule(lr){2-5} \cmidrule(lr){6-9}
    \textbf{Planner} & Succ.\(~\uparrow\) & Time\,s\(~\downarrow\) & Cost\(~\downarrow\) & Nodes\(~\downarrow\)
    & Succ.\(~\uparrow\) & Time\,s\(~\downarrow\) & Cost\(~\downarrow\) & Nodes\(~\downarrow\) \\
    \midrule
    \rowcolor{gray!10}
    \methodname (ours) & 40/50 & \rkfirst{0.72 (2.31)} & 330.6 & 2\,578 & 20/50 & \rksecond{1.66 (2.34)} & 0.082 & 1\,554 \\
    \rowcolor{gray!10}
    KPIECE (ours) & 42/50 & \rksecond{0.97 (1.72)} & 349.0 & 3\,336 & 21/50 & \rkfirst{1.51 (2.52)} & 0.108 & 1\,295 \\
    \rowcolor{gray!10}
    \methodname + polish (ours) & \rksecond{47/50} & 2.27 (3.77) & 279.9 & 1\,795 & 22/50 & 2.09 (3.34) & 0.085 & 1\,036 \\
    \rowcolor{gray!10}
    KPIECE + polish (ours) & \rkfirst{49/50} & 3.39 (4.96) & 273.4 & 2\,819 & \rkfirst{25/50} & 2.08 (5.63) & 0.094 & 1\,027 \\
    \midrule
    MCTS~\cite{worldplanner} & 31/50 & 3.05 (3.08) & 400.6 & 972 & \rksecond{24/50} & 4.75 (6.44) & 0.120 & 752 \\
    \cem~\cite{dinowm} & 46/50 & 13.33 (15.87) & 272.8 & 45\,000 & 23/50 & 28.69 (37.34) & 0.101 & 90\,000 \\
    \mppi & 45/50 & 13.70 (17.77) & 302.9 & 45\,000 & 10/50 & 36.88 (39.81) & 0.108 & 112\,500 \\
    \bottomrule
  \end{tabular}
\end{table*}

\subsection{Simulation Experiments}
\label{sec:exp:sim}

\subsubsection{Environments}
We benchmark across four simulated domains (\cref{fig:sim-envs}) designed to test distinct dynamical regimes:
\begin{itemize}
  \item \texttt{point\_maze}: A point robot navigating a planar maze (\(H=10\)) with ground-truth state proprioception \((x, y, v_x, v_y)\).
  \item \texttt{wall}: A point robot navigating through a narrow doorway in continuous space (\(H=5\)).
  \item \texttt{pusht}: The standard T-block pushing benchmark (\(H=5\)), where proprioception provides only end-effector state, leaving the objective partially blind to object motion.
  \item \texttt{rope}: Planar manipulation of a deformable cable toward a target shape (\(H=5\)).
\end{itemize}

While DINO-WM introduced a rope domain, it lacked evaluable planning tasks. We reimplemented it in MuJoCo using a composite body driven by a mocap pusher. We then collected \(5{,}000\) demonstration episodes with this environment to train our world-model checkpoint, and constructed screened task sets following our standard evaluation protocol.

\subsubsection{Baselines}
We evaluate our tree search algorithms against three sampling baselines sharing the identical world-model checkpoint and execution loop:
\begin{itemize}
  \item \textbf{CEM}: The standard Cross-Entropy Method baseline provided by the DINO-WM framework~\cite{dinowm}.
  \item \textbf{MPPI}: An internal implementation of Model Predictive Path Integral control~\cite{williams2015information} that replaces CEM's elite-selection update with an exponential-tilt weighting over the sampled rollouts, using the identical action prior and evaluation harness.
  \item \textbf{MCTS}: A re-implementation of the WorldPlanner Monte Carlo Tree Search framework~\cite{worldplanner}.
\end{itemize}
\cref{fig:sim-trees} depicts the spread of \cem, \mppi, MCTS, RRT, and KPIECE on a single \texttt{point\_maze} problem. CEM is slightly more concentrated than MPPI, and RRT covers more space with a denser tree than MCTS and KPIECE. Additionally, all CEM and MPPI trials are executed with a horizon of 5 action steps.
Sampling-based MPC is sensitive to parameters~\cite{covompc}; we sweep parameters and report the best setting.

\subsubsection{Simulation Results}
As shown in \cref{tab:main}, tree-based planners consistently match or exceed the success rates of shooting-based optimizers while operating at a fraction of the computational footprint. In contact-rich and under-actuated tasks like \texttt{pusht}, \methodname{}\,+\,polish and KPIECE\,+\,polish achieve higher task completion rates than \cem{} and \mppi{} while requiring roughly six times less wall-clock time and an order of magnitude fewer world-model propagations.

Across all domains, attaching the CEM trajectory polish provides a consistent performance boost over unpolished RRT, trading a minor fraction of wall-clock time to refine discrete tree paths into continuously optimized control sequences. KPIECE demonstrates superior state-space exploration efficiency over standard RRT; by partitioning the projected latent space, it maintains competitive task success using a sparser tree structure. We note that due to reporting planning times, path costs, and tree sizes of succcessful trials only, 

the increased path cost from KPIECE originates from the fact that higher-performing planners successfully solve harder, longer-horizon problem instances that artificially inflate average path lengths and node counts relative to earlier-failing baselines.

\subsubsection{CEM Budget}
\label{sec:exp:budget}
Shooting-based methods like \cem{} and \mppi{} are inherently tied to a fixed search budget: the total number of world-model evaluations (samples \(\times\) horizon \(\times\) optimization steps) is predetermined. Because they optimize fixed-horizon trajectories across fixed iterations, these methods must complete all assigned steps regardless of whether a valid goal-reaching plan is discovered early. In contrast, tree-based planners adjust search dynamically and can terminate immediately upon finding the first valid solution. While trees operate under an upper node limit, they typically exit well before reaching it once a path is found. Due to the mismatch between reaching the latent goal and a ground truth success, the tree methods sample a minimum number of nodes to prevent instantly returning a null tree.

\begin{table}[t]
  \centering
  \caption{\textbf{Ablation: \cem{} budget.} Performance across five settings of sample size \(\times\) optimization steps. Budget indicates world-model state evaluations per replan.}
  \label{tab:budget}
  \footnotesize
  \setlength{\tabcolsep}{5pt}
  \begin{tabular}{@{}ll rrrrr@{}}
    \toprule
    & & \multicolumn{5}{c}{\textbf{\cem{} (Samples \(\times\) Iters)}} \\
    \cmidrule{3-7}
    \textbf{Env} & \textbf{Stat} & \textbf{\(30{\times}5\)} & \textbf{\(30{\times}10\)} & \textbf{\(100{\times}5\)} & \textbf{\(300{\times}10\)} & \textbf{\(300{\times}30\)} \\
    \midrule
    \multicolumn{2}{@{}l}{\textbf{Budget}} & 750 & 1\,500 & 2\,500 & 15\,000 & 45\,000 \\
    \midrule
    \multirow{2}{*}{point\_maze}
    & Succ.  & 24/50 & 19/50 & 32/50 & 48/50 & \textbf{50/50}\rlap{\(^\ast\)} \\
    & Med\,s & 1.35  & 2.98  & \textbf{0.98} & 4.53 & 16.07 \\
    \midrule
    \multirow{2}{*}{wall}
    & Succ.  & 2/50  & 11/50 & 41/50 & \textbf{44/50}\rlap{\(^\ast\)} & \textbf{44/50} \\
    & Med\,s & 0.89  & 1.71  & \textbf{0.40} & 1.61  & 7.37 \\
    \midrule
    \multirow{2}{*}{pusht}
    & Succ.  & 0/50  & 0/50  & 21/50 & 43/50 & \textbf{46/50}\rlap{\(^\ast\)} \\
    & Med\,s & --    & --    & \textbf{1.91} & 5.41 & 13.33 \\
    \midrule
    \multirow{2}{*}{rope}
    & Succ.  & 1/50  & 2/50  & 3/50  & 17/50 & \textbf{23/50}\rlap{\(^\ast\)} \\
    & Med\,s & 6.28  & \textbf{4.23}  & 10.14 & 9.78  & 28.69 \\
    \bottomrule
  \end{tabular}

  \vspace{2pt}
  {\footnotesize \(^\ast\) Denotes default budget used in the main results (\cref{tab:main}). All cells evaluate the world model in \texttt{fp16} (\cref{tab:precision}).}
\end{table}

\cref{tab:budget} sweeps the \cem{} budget across varying parameters to contextualize the configurations used in \cref{tab:main}. At \(2{,}500\) evaluations per replan, the row whose budget is closest to the total node count our methods insert over an entire problem, \cem{} performs significantly worse, with only three rope successes. Recovering competitive accuracy requires \(15{,}000\) to \(45{,}000\) evaluations, reflecting the settings shown in \cref{tab:main}.

\subsubsection{Model Quantization}
\label{sec:exp:precision}

To justify accelerating the world model, we evaluate the impact of reduced precision on planning performance (\cref{tab:precision}). Using \texttt{fp16} precision for the world-model forward passes yields no statistically significant change in task success rates compared to \texttt{fp32}, while providing a runtime reduction (costing \(3.4\times\) less wall-clock time for \cem{} on \texttt{pusht}). All methods in \cref{tab:main} run in \texttt{fp16}.

\begin{table}[t]
  \centering
  \caption{\textbf{Ablation: model quantization.} Evaluating the world model on \texttt{pusht} (\(N{=}50\)) across different numerical representations. Relative err. refers to the error compared to the fp32 model.}
  \label{tab:precision}
  \footnotesize
  \begin{tabular}{@{}l cc c@{}}
    \toprule
    \textbf{Precision} & \textbf{Rel. Error} & \textbf{\cem\ succ.} & \textbf{\cem\ med.\,s} \\
    \midrule
    fp32 & ---                  & 45/50 & 45.58 \\
    tf32 & \(1.67\times10^{-3}\)  & 44/50 & 33.85 \\
    fp16 & \(2.54\times10^{-3}\)  & \textbf{46/50} & \textbf{13.33} \\
    \bottomrule
  \end{tabular}
\end{table}

\begin{figure*}[t]
  \refstepcounter{figure}\label{fig:live-main}
  \addtocounter{figure}{-1} %

  \begin{subfigure}[b]{1\linewidth}
    {\centering
    \includegraphics[width=\textwidth]{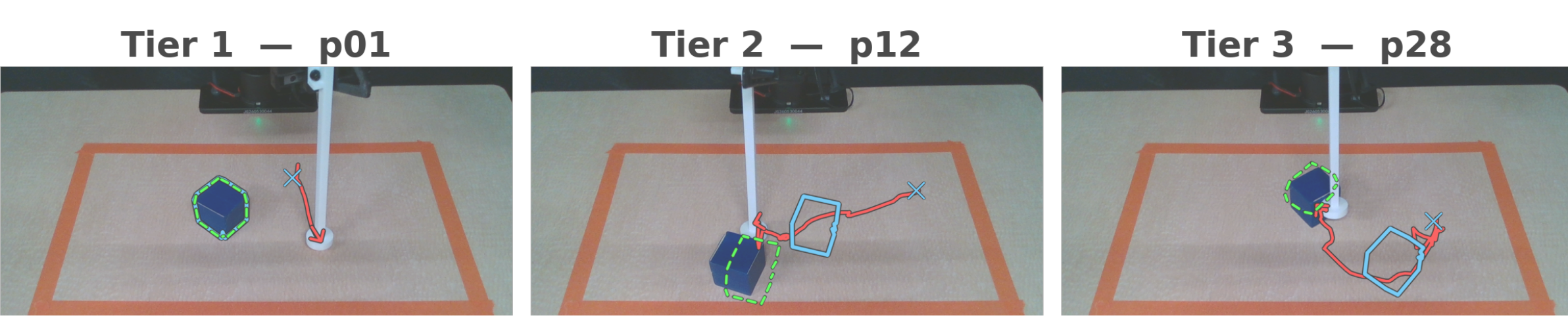}\par}
    
    {\footnotesize \linespread{0.95}\selectfont \noindent \textbf{Fig.~\thefigure(a) Yam block push difficulty tiers} on the real arm. These are solution images of real trials, overlaid with the initial block and end effector pose (blue outline and \(\times\)), end effector trajectory (red), and block goal outline (green). \textbf{Tier 1} does not involve contact with the block, while \textbf{tier 2} can be solved by pushing in a simple linear motion and \textbf{tier 3} requires complex rerouting.\par}
    
    \label{fig:live-tiers}
  \end{subfigure}

  \vspace{4pt}

  \begin{subfigure}[b]{1\linewidth}
    {\centering
    \includegraphics[width=\textwidth]{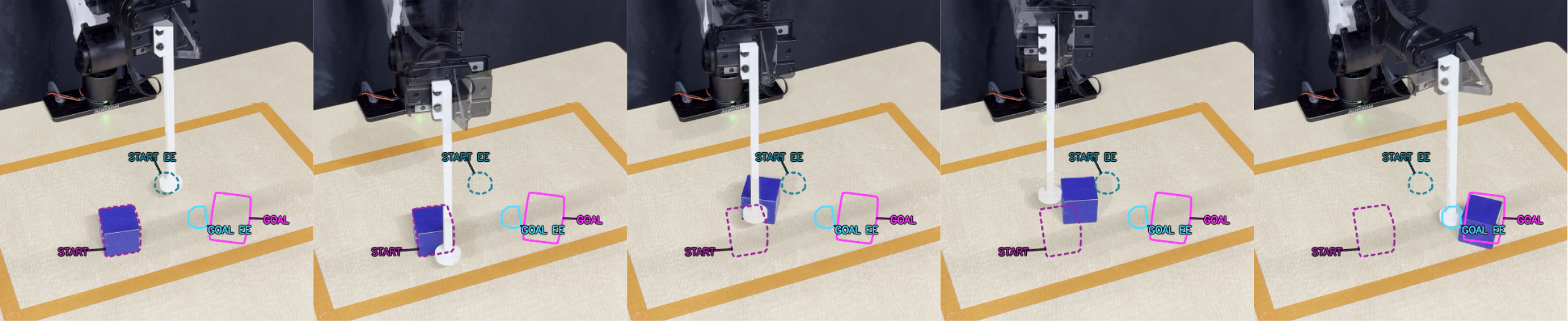}\par}

    {\footnotesize \linespread{0.95}\selectfont \noindent \textbf{Fig.~\thefigure(b) Tier-3 problem execution frame-by-frame} (Problem 27, \methodname +\,\,polish). Left to right: The end-effector starts opposite the push vector. It first traverses around the block, keeping the block stationary and leaving the cost flat for \(\sim 60\%\) of the trajectory. Only upon reaching the far side does contact occur, pushing the block into the target zone. CEM and MPPI baselines frequently stall on this initial plateau.\par}
    
    \label{fig:live-frames}
  \end{subfigure}
  \vspace{-10pt}
\end{figure*}

\subsection{Real-Robot Experiments}
\label{sec:exp:live}

We deploy our planner on a physical YAM robot arm tasked with pushing a \(50\)\,mm cube across a tabletop workspace. The underlying task formulation operates in a 4-DoF planar state space comprising the \((x, y)\) positions of both the end-effector and the target block. While block orientation is not explicitly evaluated in the ground-truth goal check, visual variations in rotation subtly inform the latent dynamics.

Perception is provided by a single overhead RGB-D camera. Visual RGB frames are fed directly to the visual world model without state estimation, while the depth stream is reserved exclusively for ground-truth block tracking during evaluation. The world model is trained on approximately \(150{,}000\) transitions collected autonomously (\(\sim 8\) hours of operation). Planning runs on a dedicated NVIDIA RTX 4090 GPU, streaming planned actions to a separate motion controller on the arm. The planning objective minimizes distance in latent space via a spatial soft-argmax head targeting the block position. A trial is marked successful if the final frame's ground-truth block error is within \(25\)\,mm (half block width) and the end-effector does not stray past an acceptable operational boundary.

\begin{figure}[b]
  \centering
  \includegraphics[width=\columnwidth]{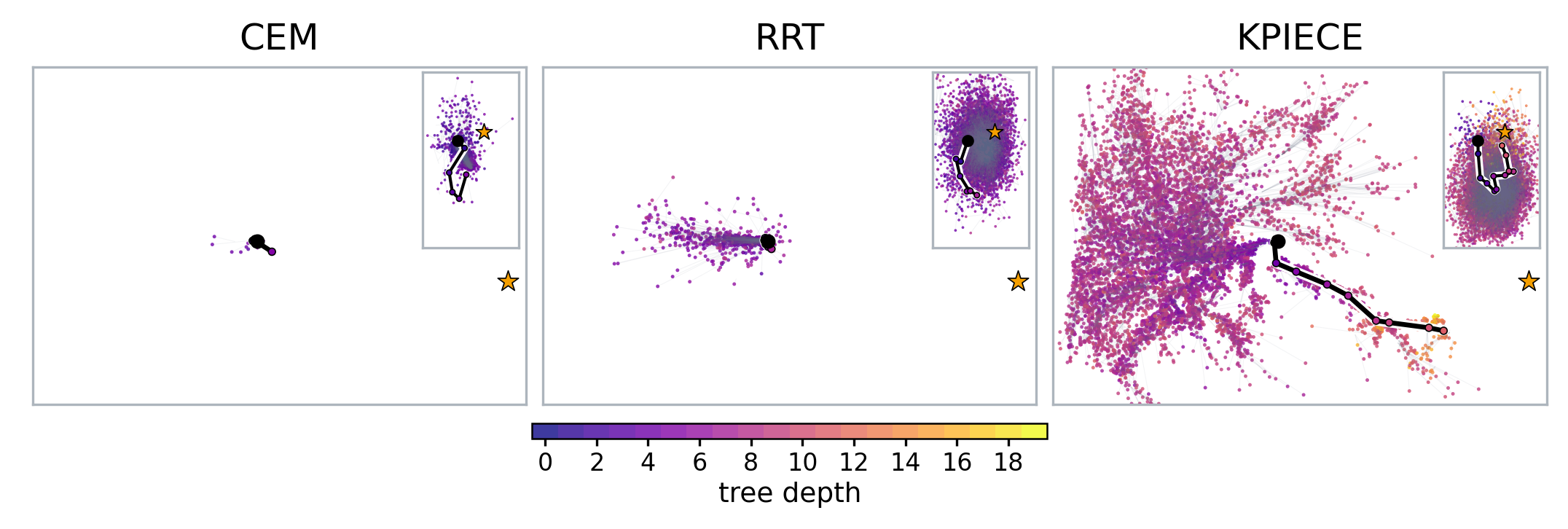}
  \caption{\textbf{Ablation: case study.} Visualization of the first search tree in P27 for CEM, RRT, and KPIECE, shown in the end-effector's (EE) XY position (top right corner) and the block's XY position (main plot). Planned fixed-horizon trajectory for EE and prediction of block motions are plotted as lines, while the lighter color indicates a deeper node. The star indicates the goal position.}
  \label{fig:livtree}
\end{figure}

\begin{table}[t]
  \centering
  \caption{\textbf{Real-robot results.} YAM arm block pushing (\(30\) total problems across \(3\) tiers). \emph{Succ.} indicates trials reaching \(<25\)\,mm final block distance. \emph{Repl.} is the median MPC replans, \emph{Plan\,s} is the median total planning time per problem, and \emph{Err.} is the mean final block error on successful trials.}
  \label{tab:real}
  \footnotesize
  \setlength{\tabcolsep}{4pt}
  \begin{tabular}{@{}l c rrrr@{}}
    \toprule
    \textbf{Tier} & \textbf{Planner} & \textbf{Succ.} & \textbf{Repl.} & \textbf{Plan\,s} & \textbf{Err.\,mm} \\
    \midrule
    \multirow{4}{*}{1}
    & \methodname + polish (ours) & \rkfirst{10/10} & 2.0 & 10.2 & 5.1 \\
    & KPIECE + polish (ours)      & \rkfirst{10/10} & \rkfirst{1.0} & \rkfirst{5.8} & \rkfirst{3.3} \\
    & \cem                         & \rkfirst{10/10} & \rkfirst{1.0} & 9.4  & 6.0 \\
    & \mppi                        & \rkfirst{10/10} & \rkfirst{1.0} & \rksecond{6.0} & \rksecond{4.6} \\
    \midrule
    \multirow{4}{*}{2}
    & \methodname + polish (ours) & \rkfirst{10/10} & \rksecond{2.0} & \rksecond{10.2} & \rkfirst{12.2} \\
    & KPIECE + polish (ours)      & \rkfirst{10/10} & 4.0 & 24.9 & \rksecond{13.1} \\
    & \cem                         & \rksecond{8/10} & \rkfirst{1.5} & \rkfirst{9.4} & \rkfirst{12.2} \\
    & \mppi                        & \rksecond{8/10} & 3.0 & 18.0 & 13.6 \\
    \midrule
    \multirow{4}{*}{3}
    & \methodname + polish (ours) & \rkfirst{9/10} & \rksecond{5.0} & \rksecond{30.3} & 15.0 \\
    & KPIECE + polish (ours)      & \rksecond{8/10} & \rkfirst{4.0} & \rkfirst{27.7} & 14.1 \\
    & \cem                         & 5/10  & 7.5 & 46.9 & \rksecond{12.8} \\
    & \mppi                        & 5/10  & 7.0 & 42.0 & \rkfirst{11.1} \\
    \midrule
    \multirow{4}{*}{\textbf{Total}}
    & \methodname + polish (ours) & \rkfirst{29/30} & 2.5 & \rksecond{14.6} & 10.6 \\
    & KPIECE + polish (ours)      & \rksecond{28/30} & 3.0 & 18.8 & 9.9 \\
    & \cem                         & 23/30 & \rkfirst{2.0} & \rkfirst{12.5} & \rksecond{9.6} \\
    & \mppi                        & 23/30 & \rkfirst{2.0} & 15.0 & \rkfirst{9.1} \\
    \bottomrule
  \end{tabular}
\end{table}

\subsubsection{Problem Tiers}
We define \(30\) evaluation problems across \(10\) unique initial configurations, categorized into three distinct difficulty tiers based on task structure (\cref{fig:live-tiers}a):
\begin{enumerate}[leftmargin=*, itemsep=2pt, topsep=3pt]
  \item \textbf{Tier 1 (Reposition):} Reach an end-effector target pose without displacing the block (no contact required).
  \item \textbf{Tier 2 (Straight Push):} Perform a short, direct push to move the block to a nearby target.
  \item \textbf{Tier 3 (Complex Push):} Require non-prehensile long-horizon navigation: the end-effector must move around the block to its far side before pushing it back toward the target goal.
\end{enumerate}

\subsubsection{Live Results}
\cref{tab:real} details performance across tiers. In Tier 1, all methods succeed with minor variations in execution speed. Performance diverges on Tier 2, where both tree planners achieve success on the full set compared to \(8/10\) for shooting-based baselines.

The gap widens on Tier 3 tasks requiring detour planning. \methodname\,+\,polish completes \(9/10\) and KPIECE\,+\,polish \(8/10\), compared to \(5/10\) for CEM and MPPI. On these tasks, shooting-based methods incur higher total planning times and require more replanning steps. Because unsuccessful baseline runs fail to establish contact, they repeatedly exhaust their iteration limits on uninformative plateaus.

\subsubsection{Case Study: Search Trees on Problem 27}

To better understand the exploration strategies of different planners, we run a case study to visualize the expanded tree for a single planning query in the real-world push block experiment. We choose a Tier-3 problem, P27 shown in \cref{fig:live-frames}b, and visualize the tree from the first iteration in \cref{fig:livtree}. Intuitively, the CEM spends most of its computational budget in the same regions, showing a dense tree in an underexplored space. On the other hand, because KPIECE's selection strategy encourages higher coverage in the space, it grows a tree that explores the space evenly with deeper branches. RRT offers a balance between them with a reasonably dense tree and some exploration. KPIECE trades broader coverage for deeper branches, which we believe explains why, in \cref{tab:real}, KPIECE spends more iterations correcting itself from compounding world-model errors incurred by the longer plan.

%% file: sections/07-conclusion.tex
\section{Conclusion}
\label{sec:conc}

We have presented a kinodynamic sampling-based planner that uses a frozen visual world model as its propagator, with RRT-style nearest-neighbor and KPIECE-style projection-based exploration of the model's latent space, batched expansion, and a short CEM refinement of the returned plan.
In simulation and on a real block-pushing task, these exploration strategies are fast and effective, and they help most on problems where the robot must first move away from immediate reward.
Limitations remain; the goal metric biases search and is the weakest link in the pipeline.
Much as in reward shaping, our planners could exploit other goal representations in the latent space.
Performance could also improve with a distilled or lower-resolution propagator, which would allow faster and deeper search since the predictor is the dominant cost of every expansion.

%% file: sections/10-ack.tex
\section{Acknowledgment}
\label{sec:ack}
ChatGPT/Codex and Claude code were used for language polishing and editorial assistance. All technical content, equations, experimental results, and final manuscript text were reviewed and verified by the authors.